\pdfoutput=1
\documentclass[11pt]{article}

\usepackage{ACL2023}

\usepackage{times}
\usepackage{latexsym}
\usepackage{amsmath}
\usepackage{txfonts}
\usepackage{float}
\newcommand{\macrof}{$\mathrm{m}$-$\mathrm{F_1}$\xspace}
\newcommand{\microf}{$\mathrm{\muup}$-$\mathrm{F_1}$\xspace}
\usepackage{graphicx}
\usepackage{multirow}
\usepackage{subcaption}
\usepackage{enumitem}
\usepackage{booktabs}
\usepackage[T1]{fontenc}

\usepackage{xspace}

\usepackage[utf8]{inputenc}

\usepackage{microtype}

\usepackage{inconsolata}

\title{Efficient Document Embeddings via \\ Self-Contrastive  Bregman Divergence Learning}

\author{Daniel Saggau\thanks{\hspace{0.5em}The authors contributed equally to this work.} \qquad Mina Rezaei$^*$ \qquad Bernd Bischl \\
Department of Statistics, Ludwig Maximilian University of Munich (LMU), Germany \\ 
Munich Center for Machine Learning (MCML), Germany \\
\AND
  Ilias Chalkidis \\
  Department of Computer Science, University of Copenhagen, Denmark
  }

\begin{document}
\maketitle

\begin{abstract}

Learning quality document embeddings is a fundamental problem in natural language processing (NLP), information retrieval (IR), recommendation systems, and search engines. Despite recent advances in the development of transformer-based models that produce sentence embeddings with self-contrastive learning, the encoding of long documents (Ks of words) is still challenging with respect to both efficiency and quality considerations. Therefore, we train Longfomer-based document encoders using a state-of-the-art unsupervised contrastive learning method (SimCSE). Further on, we complement the baseline method -siamese neural network- with additional convex neural networks based on functional Bregman divergence aiming to enhance the quality of the output document representations.
We show that overall the combination of a self-contrastive siamese network and our proposed neural Bregman network outperforms the baselines in two linear classification settings on three long document topic classification tasks from the legal and biomedical domains.

\end{abstract}
\section{Introduction}

The development of quality document encoders is of paramount importance for several NLP applications, such as long document classification tasks with biomedical~\cite{johnson2016mimic}, or legal \cite{chalkidis-etal-2022-lexglue} documents, as well as information retrieval tasks~\cite{,chalkidis-etal-2021-regulatory,coliee2021,bioasq-2022}. Despite the recent advances in the development of transformer-based sentence encoders~\citep{reimers2019sentence, gao_simcse_2021,liu-etal-2021-fast,klein2022micse} via unsupervised contrastive learning, little do we know about the potential of neural document-level encoders targeting the encoding of long documents (Ks of words).

\begin{table}[]
    \centering
    \begin{tabular}{l|r}
        \toprule
         \bf Training Corpus & \bf Average Text Length \\
         \midrule
         \multicolumn{2}{c}{\citet{reimers2019sentence} \emph{inter alia}} \\
         \midrule
         SNLI & 22 \\
         MNLI & 113 \\
         MS Marco & 335 \\
         Wikipedia &  200 \\
         \midrule
         \multicolumn{2}{c}{Our Work} \\
         \midrule
         ECtHR & 1,613 \\
         MIMIC & 1,621\\
         SCOTUS & 5,853\\
         \bottomrule
    \end{tabular}
    \vspace{-2mm}
    \caption{Text length across corpora that have been used for self-contrastive pre-training in the NLP literature.}
    \label{tab:length}
    \vspace{-5mm}
\end{table}

The computational complexity of standard Transformer-based models~\cite{vaswani2017attention, devlin2019bert} (PLMs) given the quadratic self-attention operations poses challenges in encoding long documents. To address this computational problem, researchers have introduced efficient sparse attention networks, such as Longformer~\cite{beltagy2020longformer}, BigBird~\cite{zaheer2020big}, and Hierarchical Transformers~\cite{chalkidis-etal-2022-hat}. Nonetheless, fine-tuning such models in downstream tasks is computationally expensive; hence we need to develop efficient document encoders that produce quality document representations that can be used for downstream tasks out-of-the-box, i.e., without fully (end-to-end) fine-tuning the pre-trained encoder, if not at all.

Besides computational complexity, building good representation models for encoding long documents can be challenging due to document length. Long documents contain more information than shorter documents, making it more difficult to capture all the relevant information in a fixed-size representation. In addition, long documents may have sections with different topics, which increases the complexity of encoding that usually leads to collapsing representations~\cite{jing2022understanding}. Moreover, long documents can be semantically incoherent, meaning that content may not be logically related or may contain irrelevant information. For these reasons, it is challenging to create a quality representation that captures the most important information in the document.

To the best of our knowledge, we are the first to explore the application of self-contrastive learning for long documents (Table~\ref{tab:length}). 
The contributions of our work are threefold:\vspace{2mm} 

\noindent \hspace{1mm}(i) We train Longfomer-based document encoders using a state-of-the-art self-contrastive learning method, SimCSE by \citet{gao_simcse_2021}.\vspace{2mm}   

\noindent \hspace{1mm}(ii) We further enhance the quality of the latent representations using convex neural networks based on functional Bregman divergence. The network is optimized based on self-contrastive loss with divergence loss functions~\cite{rezaei2021deep}.\vspace{2mm}  

\noindent \hspace{1mm} (iii) We perform extensive experiments to highlight the empirical benefits of learning representation using unsupervised contrastive and our proposed enhanced self-contrastive divergence loss. We compare our method with baselines on three long document topic classification tasks from the legal and biomedical domain.

\section{Related Work}

\paragraph{Document Encoders}
The need for quality document representations has always been an active topic of NLP research. 
Initial work on statistical NLP focused on  representing documents as Bag of Words (BoW), in which direction TF-IDF representations were the standard for a long time.
In the early days of deep learning in NLP, models developed to represent words with latent representations, such as Word2Vec~\cite{Mikolov2013}, and GloVe~\cite{Pennington2014}.
Within this research domain, the use of word embedding centroids as document embeddings, and the development of the Doc2Vec~\cite{le2014distributed} model were proposed.
Given the advanced compute needs to encode documents with neural networks, follow-up work mainly developed around sentence/paragraph-level representations, such as Skip Thoughts of \citet{kiros2015skip}, which relies on an RNN encoder. 
In the era of pre-trained Transformer-based language models, \citet{reimers2019sentence} proposed the Sentence Transformers framework in order to develop quality dense sentence representations. Many works followed a similar direction relying on a self-supervised contrastive learning setup, where most ideas are adopted mainly from Computer Vision literature~\cite{chen2020simple,bardes2022vicreg}.

\paragraph{Self-Supervised Contrastive Learning in NLP}

Several self-contrastive methods have been proposed so far for NLP applications. To name a few: MirrorRoBERTa~\cite{liu-etal-2021-fast}, SCD~\cite{kleinscd}, miCSE~\cite{klein2022micse}, DeCluTR~\cite{giorgi2021declutr}, and SimCSE~\cite{gao_simcse_2021} -- described in Section~\ref{sec:simcse}--, all create augmented versions (views) of the original sentences using varying dropout and comparing their similarity. The application of such methods is limited to short sentences and relevant downstream tasks, e.g., sentence similarity, while these methods do not use any additional component to maximize diversity in latent feature representations.

\section{Methods}
\subsection{Base Model - Longformer} We experiment with Longformer~\citep{beltagy2020longformer}, a well-known and relatively simple sparse-attention Transformer. 
Longformer uses two sets of attention, namely sliding window attention and global attention.
Instead of using the full attention mechanism, the sliding-window attention gives local context higher importance. 
Given a fixed window size $w$, each token attends to $\frac{1}{2}w$ tokens on the respective side.
The required memory for this is $O(n \times w)$.
Sliding-window attention is combined with global attention from/to the \texttt{[CLS]} token.\vspace{2mm}

\noindent\textbf{Domain-Adapted Longformer:} As a baseline, we use $\text{Longformer}_{\text{DA}}$ models which are Longformer models warm-started from domain-specific PLMs.
To do so, we clone the original positional embeddings $8\times$ to encode sequences up to 4096 tokens. 
The rest of the parameters (word embeddings, transformers layers) can be directly transferred, with the exception of Longformer's global attention K, Q, V matrices, which we warm-start from the standard (local) attention matrices, following~\citet{beltagy2020longformer}.
All parameters are updated during training.

For legal applications (Section~\ref{sec:datasets}), we warm-start our models from Legal-BERT~\cite{chalkidis-etal-2020-legal}, a BERT model pre-trained on diverse English legal corpora, while for the biomedical one, we use BioBERT ~\cite{lee2020biobert}, a BERT model pre-trained on biomedical corpora.\vspace{2mm}

\begin{figure}[t]
 \centering
   \includegraphics[width=\columnwidth]{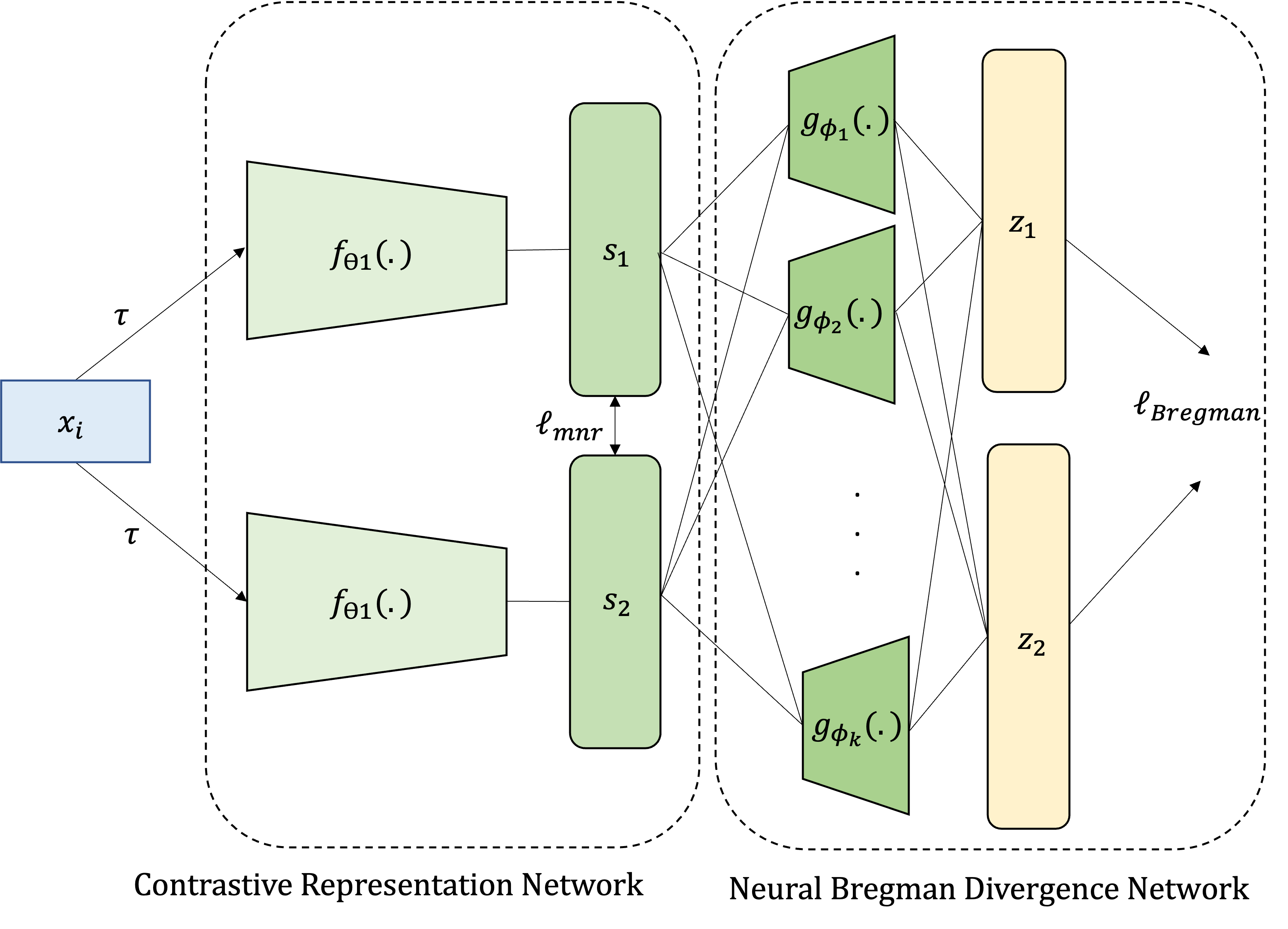}%
   \vspace{-3mm}
   \caption{Illustration of our proposed self-contrastive method combining SimCSE of \citet{gao_simcse_2021} (left part) with the additional Bregman divergence networks and objective of \citet{rezaei2021deep} (right part).}
   \label{fig:network}
\end{figure}

\subsection{Self-supervised Contrastive Learning}
\label{sec:simcse}
To use our $\text{Longformer}_{\text{DA}}$ for self-supervised contrastive learning, we need to use a Siamese network architecture (left part of Figure~\ref{fig:network}).
Assume we have mini-batch $\mathcal{D}=\left\{\left(x_i\right)\right\}_{i=1}^N$ of $N$ documents.
As positive pairs $(x_i,{x_i}^+)$, the method uses augmented (noised) versions of the input feature $x_i$.
As negative pairs $(x_i,{x_i}^-)$, all remaining N-1 documents in a mini-batch are used.
The augmentations take place in the encoder block $f_{\theta}$  of the model.
$\theta$ is the parameterization of the encoder.
We use the SimCSE~\cite{gao_simcse_2021} framework, in which case the encoder $f_{\theta}$ is a pre-trained language model, $\text{Longformer}_{\text{DA}}$ in our case, and augmentation comes in the form of varying token dropout (masking) rate ($\tau$).
The loss objective used in the unsupervised version of SimCSE is the multiple negatives ranking loss ($\ell_\text{mnr}$):

\begin{equation}
\ell_\text{mnr}=-\frac{1}{n} \sum_{i=1}^n \frac{\exp \left(f\left(s_i, \tilde{s}_i\right)\right)}{\sum_j \exp \left(f\left(s_i, s_j\right)\right)}
\end{equation}

\noindent where $\tilde{s_i}$ is the positive augmented input sequence in the mini-batch, and $\tilde{s_j}$ are the negatives.
Multiple negatives ranking loss takes a pair of representations ($s_i$, $\tilde{s_i}$) and compares these with negative samples in a mini-batch.
In our experiments, we train such models, dubbed $\text{Longformer}_{\text{DA+SimCSE}}$.

\begin{table*}[t]
\centering
\scalebox{.9}{
\begin{tabular}{lcc|cc|cc|c|cc}
\toprule
  \multirow{2}{*}{\bf Method}    & \multicolumn{2}{c}{\bf ECtHR} & \multicolumn{2}{c}{\bf SCOTUS} & \multicolumn{2}{c}{\bf MIMIC}  & \bf Avg. & \multicolumn{2}{c}{\bf Training Efficiency}\\
    & \microf & \macrof & \microf & \macrof & \microf & \macrof & \microf & Time (h) & Params (\%) \\
    \midrule
    \multicolumn{7}{l}{\bf \textsc{Document Embedding + MLP}} \\     
    \midrule 
    $\text{Longformer}_{\text{DA}}$ & 61.4 & 47.8 & 65.7  & 50.5 & 63.9 &  48.3 & 63.6  & 4.5h  & 0.5\% \\
  $\text{Longformer}_{\text{DA+SimCSE}}$ & \underline{64.4} & \underline{55.0} & \underline{69.2} & \underline{57.5} &  \underline{66.0} &  \textbf{52.9} & \underline{66.5} & >> & >>\\
$\text{Longformer}_{\text{DA+SimCSE+Bregman}}$ & \textbf{64.8} & \textbf{56.3}  & \textbf{69.7} & \textbf{58.8} & \textbf{66.7} & \underline{51.7} & \bf 67.1 & >> & >> \\
    \midrule
    \multicolumn{7}{l}{\bf \textsc{Document Embedding + Linear Layer}} \\     
    \midrule   
   $\text{Longformer}_{\text{DA}}$ & \textbf{73.7} & \textbf{62.4} & 69.3 & 59.0  & \underline{59.4} & 21.7 & \underline{67.5} & 1h & 0.5\% \\   
    $\text{Longformer}_{\text{DA+SimCSE}}$ & 70.6 & 56.2 & \underline{69.6} & \underline{60.9} & 59.2 & \textbf{23.0} & 66.5 & >> & >> \\ 
    $\text{Longformer}_{\text{DA+SimCSE+Bregman}}$ & \underline{73.3} & \underline{59.5} & \textbf{71.4} & \textbf{62.0} & \textbf{59.6} & \underline{22.7} & \bf 68.1 & >> & >> \\
\midrule
\multicolumn{7}{l}{\bf \textsc{End-to-End Fine-tuning (Ceiling)}} \\     
    \midrule  
  $\text{Longformer}_{\text{DA}}$  & 78.8  &  71.5 & 75.2 & 63.2 & 78.9 & 56.4 & 77.6 & 8h  & 100\% \\
\bottomrule
\end{tabular}
}
\caption{Test Results for all methods across all datasets. Best performance in \textbf{bold}, and second-best score is \underline{underlined}. We also report average training time and the percentage of parameters that are trainable.} 
\label{tab:ablation_summary}
\end{table*}

\subsubsection{Bregman Divergence Loss}

We complement this method with an additional ensemble of subnetworks optimized by functional Bregman divergence aiming to improve the output document latent representations further. Specifically, the embedding of self-contrastive networks further passes to $k$-independent subnetworks to promote diversity in feature representations.

\noindent The $s_i$ and $s_j$ vectors from the contrastive framework are mapped to $k$-independent ensemble of neural networks that are optimized using functional Bregman divergence.
\begin{equation}
\begin{aligned}
G_\phi(s_{a}, s_{b})=\phi(s_{a})-\phi(s_{b})- \\ 
\int[s_{a}(x)-s_{b}(x)] \delta \phi(s_{b})(x) d \mu(x)
\end{aligned}
\end{equation}

\noindent$s_{a}$ and $s_{b}$ are vectors output by the self-contrastive network, and $\phi$ is a strictly convex function and can be described via a linear functional, consisting of weights $w_k$ and biases $\epsilon_k$.
The function $\phi(s_a)$ is approximate by:
\begin{equation}
\phi(s_{a})=\underset{\left(w, \epsilon_w\right) \in Q}{s u p} \int s_{a}(x) w(x) d x+\epsilon_w
\end{equation}

\noindent We take the empirical distribution of the projection representation to compute $\hat{s_a}$ and $\hat{s_b}$.
Specifically we define: $\hat{s_i}\!=\!\operatorname{argmax}_k[\int s_a(x) w_k(x) d x+\epsilon_{k}]$ for i = (a,b).
Using the above specification and $\phi(s_a)$, we get the following functional divergence term: 
\begin{equation}
\begin{aligned}
G(s_a, s_b)= 
(\int s_a(x) w_{\hat{s_a}}(x) d x+\epsilon_{\hat{s_a})}- \\ (\int s_a(x) w_{\hat{s_b}}(x) d x+\epsilon_{\hat{s_b}})
\end{aligned}
\end{equation}

\noindent Each sub-network produces a separate output (right part of Figure~\ref{fig:network}). 
The divergence is then computed using the output at point $\hat{s_a}$ and $\hat{s_b}$ using the projections as input.
We convert the divergence to similarity using a Gaussian kernel as done by \citet{rezaei2021deep}.\footnote{\citet{rezaei2021deep} explore various monotone transformations.
The Gaussian kernel performed best compared to other transformation methods.}

\begin{equation}
\psi=\exp \left(-G / 2 \sigma^2\right)
\end{equation}

\noindent The mini-batch has size N.
For empirical distributions $s_a\alpha(z_i),s_b(z_j)$ where i and j are the respective index for the two branches and z the projector representation, we have:
\begin{equation}
\begin{aligned}
\ell_{\operatorname{Bregman}(s_{a}(z_i), s_{b}(z_j))}=&   \frac{-\log(\exp \left(\psi_{i, j}\right) }{\sum_{t=1}^N \exp \left(\psi_{i, k}\right))} 
\end{aligned}
\end{equation}

\noindent The final objective function is computed on the combination of  as follows: 

\begin{equation}
L_\text{Total}= \ell_\text{mnr} + \lambda \cdot \ell_\text{Bregman} 
\end{equation}

\noindent Where $\lambda$ is a scalar hyperparameter to weigh the relative importance of the Bregman divergence and contrastive loss.
In our experiments, we train such models, dubbed $\text{Longformer}_{\text{DA+SimCSE+Bregman}}$.

\section{Experimental Set-up}
\subsection{Datasets and Tasks} 
\label{sec:datasets}

\noindent\textbf{ECtHR}~\cite{chalkidis-etal-2021-paragraph} dataset contains 11k cases from the European Court of Human Rights (ECtHR). This is a multi-label topic classification task, where given the facts of an ECtHR case, the model has to predict the alleged violated ECtHR article among ten such articles (labels).\vspace{2mm}

\noindent\textbf{SCOTUS}~\cite{chalkidis-etal-2022-lexglue} dataset contains 4.7k cases from the Supreme Court of US (SCOTUS). This is a single-label multi-class topic classification task, where given a SCOTUS opinion, the model has to predict the relevant area among 14 issue areas (labels).\vspace{2mm}

\noindent\textbf{MIMIC}~\cite{johnson2016mimic} dataset contains approx.\ 50k discharge summaries from US hospitals. Each summary is annotated with one or more codes (labels) from the ICD-9 hierarchy, which has 8 levels in total.  We use the 1st level of ICD-9, including 19 categories, respectively. This is a multi-label topic classification task, where given the discharge summary, the model has to predict the relevant ICD-9 top-level codes (labels).\vspace{2mm}

\subsection{Experimental Settings}
To get insights into the quality of the learned representations out-of-the-box, we train classifiers using document embeddings  as fixed (frozen) feature representations. 
We consider two linear classification settings:
(i) Linear evaluation plugging a MLP classification head on top of the document embeddings; (ii) Linear evaluation plugging a linear classifier on top of the document embeddings.

\section{Results and Discussion}

In Table~\ref{tab:ablation_summary}, we present the results for all examined Longformer  variants across the three examined datasets and two settings using macro-F1 (\macrof) and micro-F1 (\microf) scores. 
\vspace{2mm}

\noindent\textbf{Classification performance:}
In the last line of Table~\ref{tab:ablation_summary}, we present the results for the baseline $\text{Longformer}_{\text{DA}}$ model fine-tuned end-to-end, which is a `ceiling' for the expected performance, comparing to the two examined linear settings, where the document encoders are not updated. We observe that in the SCOTUS dataset training models with an MLP head are really close to the ceiling performance (approx.~1-4p.p. less in \microf). The gap is smaller for both models trained with the self-contrastive objective (+SimCSE, +SimCSE+Bregman), especially the one with the additional Bregman divergence loss, where the performance decrease in \microf is only 1 p.p.  

In the other two datasets (ECtHR and MIMIC), the performance of the linear models is still approx. 10-15 p.p. behind the ceilings in \microf. In ECtHR, we find that self-contrastive learning improves performance in the first settings by 3 p.p. in \microf, while the additional divergence Bregman loss does not really improve performance. This is not the case, in the second linear setting (second group in Table~\ref{tab:ablation_summary}), where the baseline outperforms both models. Similarly in MIMIC, we observe that self-contrastive learning improves performance in the first settings by 3 p.p. in \microf, but the performance is comparable given linear classifiers. Overall, our enhanced self-contrastive method leads to the best results compared to its counterparts.

In Table~\ref{tab:setfit_scotus}, we also present results on SCOTUS in a few-shot setting using the SetFit~\cite{tunstall-etal-2022} framework, where Bregman divergence loss improves performance compared to the baselines.\vspace{1mm}

\begin{table}[t]
\centering
\begin{tabular}{lcc}
\toprule
 {\bf Model} & \microf & \macrof  \\
 \midrule
     $\text{Longformer}_{\text{DA}}$ & \underline{54.9} & \underline{48.1}  \\  
     >> + SimCSE & 51.8 & 43.6  \\
     >> + SimCSE + Bregman & \textbf{56.9} & \textbf{48.5} \\ 
\bottomrule
\end{tabular}
\caption{Test Results for all Longformer variants for SCOTUS.  Best performance in \textbf{bold}, and second-best score is \underline{underlined}.}
\label{tab:setfit_scotus}
\end{table}

Given the overall results, we conclude that building subnetwork ensembles on top of the document embeddings can be a useful technique for encoding long documents and can help avoid the problem of collapsing representations, where the model is unable to capture all the relevant information in the input. Our approach has several advantages for long-document processing:\vspace{2mm}

\noindent\textbf{Efficiency considerations:} In Table~\ref{tab:ablation_summary}, we observe that in both linear settings where fixed document representations are used, the training time is 2-8$\times$ decreased compared to end-to-end fine-tuning, while approx.~0.5\%  of the parameters are trainable across cases, which directly affects the compute budget. We provide further information on the size of the models in Appendix \ref{sec:params}.\vspace{2mm}

\noindent\textbf{Avoidance of collapsing representations:} When processing long documents, there is a risk that the representation will collapse~\cite{jing2022understanding}, meaning that the model will not be able to capture all the relevant information in the input. By mapping the document embedding from the base encoder into smaller sub-networks, the risk of collapsing representations is reduced, as the divergence loss attempts to reduce redundancy in the feature representation by minimizing the correlation. The results shown in Table~\ref{tab:setfit_scotus} in a low-resource setting further highlight the advantage of training a Longformer with contrastive divergence learning.

\vspace{-2mm}
\section{Conclusions and Future Work}
\vspace{-2mm}
We proposed and examined self-supervised contrastive divergence learning for learning representation of long documents. Our proposed method is composed of a self-contrastive learning framework followed by an ensemble of neural networks that are optimized by functional Bregman divergence. Our method showed improvement compared to the baselines on three long document topic classifications in the legal and biomedical domains, while the improvement is more vibrant in a few-shot learning setting. In future work, we would like to further investigate the impact of the Bregman divergence loss in more classification datasets and other NLP tasks, e.g., document retrieval.

\section*{Limitations}

In this work, we focus on small and medium size models (up to 134M parameters), while recent work in Large Language Models (LLMs) targets models with billions of parameters\cite{brown2020language, Chowdhery2022}. 
It is unclear how well the performance improvement from the examined network architecture would translate to other model sizes or baseline architectures, e.g., GPT models. 

Further on, it is unclear how these findings may translate to other application domains and datasets, or impact other NLP tasks, such as document retrieval/ranking. We will investigate these directions in future work.

\section*{Acknowledgments}
Mina Rezai and Bernd Bisch were supported by the Bavarian Ministry of Economic Affairs, Regional Development and Energy through the Center for Analytics – Data – Applications (ADA-Center) within the framework of BAYERN DIGITAL II (20-3410-2-9-8).M. R. and B. B. were supported by the German Federal Ministry of Education and Research (BMBF) Munich Center for Machine Learning (MCML). This work was also partly funded by the Innovation Fund Denmark (IFD).\footnote{\url{https://innovationsfonden.dk/en}}

\bibliography{anthology,custom}
\bibliographystyle{acl_natbib}

\appendix

\section{Hyper-parameter Optimization}
\label{sec:appendix}
\noindent\textbf{Continued Pre-training:}
We define the search space based on previous studies such as \citet{rezaei2021deep} and \citet{gao_simcse_2021}.
For the contrastive Bregman divergence, we benchmark the performance for the first-stage hyper-parameters on the downstream task to tune the respective hyper-parameters. 
We use mean pooling for all settings. The learning rate, the total optimization steps, the use of a batch-norm layer, the ${\sigma}$ parameter, the number of sub-networks $g$, and the batch size are grid-searched.
Temperature (.1) and the input length to 4096 are fixed beforehand.
The learning rate for these models was 3e-5.
We run 50.000 optimization steps for each model.
\vspace{2mm}

\begin{table*}[!t]
\centering
\scalebox{.9}{
\begin{tabular}{l c c c c c c c c c c} 

\toprule
  \bf Hyper-parameters & \microf & \macrof & \microf & \macrof & \microf & \macrof & \microf & \macrof & \microf & \macrof \\
\midrule
    g $\in$ [2,5,8,10,20] & 74.1 & \underline{64.3} & \underline{74.3} & 62.2 & 72.1 & 61.0 & \textbf{75.2} & \textbf{67.7} &73.9  & 63.1 \\
    $\sigma$ $\in$ [1,1.5,2,2.5,3] & 73.3 & 63.0 & 73.6 & 61.2  & \textbf{75.2} & \textbf{67.7} & 73.0 & 62.0 &  \underline{73.9} &\underline{64.1} \\
    Steps  $\in$ [10-50k] & 74.21  & 62.79 & 74.14& 63.44 & \textbf{75.6} & \underline{63.5} & 73.5 & 63.36 & \underline{75.2}& \textbf{67.7} \\
    Batch size $\in$ [2,4,8,12] & \textbf{75.2} & \textbf{67.7} & \underline{74.36} & \underline{64.2} & 73.9 & 62.6 & 74.21 & 62.9 & - & -\\
    $\lambda$ $\in$ [.1,2,4,5,10] & \underline{75.1} & \underline{65.3} & \textbf{75.2}&  \textbf{67.7}  & 74.79 & 63.4 & 74.1& 63.7 & \textbf{75.2} & 64.0\\
\bottomrule
\end{tabular}
}
\caption{\macrof \& \microf performance benchmark for end-to-end training with SCOTUS}
\label{tab:subnets}
\end{table*}

\noindent\textbf{Training for classification tasks:} We used AdamW as an optimizer. Bayesian optimization is used to tune the hype-rparameters learning rate, number of epochs and batch size. We use mean pooling for all settings.
Early stopping is set to a patience score of 3.\footnote{We also experimented with other patience scores but experiments suggest that 3 epochs results in the best performance.}
These parameters were fixed after some early experiments.
We use a learning rate of 1e-4 and run ECTHR and SCOTUS for 20 and 30 epochs respectively for the MLP head setting. 
For MIMIC we used 10 epochs for the MLP head and had to truncate the maximum sequence length to 2048 due to computational constraints.
For each task we compared multiple different training checkpoints of our encoder.
The reported results are the best performing checkpoints.

\section{Number of parameters}
\label{sec:params}
Table \ref{tab:efficiency-performance} shows the number of parameters for the different models.
Modding the transformer to a Longformer adds 6M parameters for LegalBERT small and 24M parameters for BioBERT medium.
By working with LegalBERT-small and BioBERT-base we cover both small and medium sized models.
\begin{table}[h]
\centering
\begin{tabular}{lc}
\toprule
  \textbf{Model}    &   \#Params \\
    \midrule  
    $\text{BioBert}_{\text{Base}}$  & 109M  \\
    $\text{Longformer}_{\text{Base}}$   & 148M  \\
    $\text{LegalBERT}_{\text{small}}$ & 35M \\
 $\text{Longformer}_{\text{Legal-DA + SimCSE + Bregman}}$  & 41M \\
    $\text{Longformer}_{\text{Bio-DA}}$  & 134M \\
    $\text{Longformer}_{\text{MLP}}$ & .27M \\
\bottomrule
\end{tabular}
\caption{Number of Parameters for the Longformer variants.}
\label{tab:efficiency-performance}
\end{table}

\section{Pooling methods}

We evaluate Mean, Max and \texttt{[CLS]} pooling. 
Results for end-to-end fine-tuning can be found in the table \ref{tab:pool}.
Our results show that using mean pooling during continued pre-training in combination with max-pooling for classification could further enhance the performance instead of using the same pooling method for both stages.
\begin{table}[h]
\centering
\begin{tabular}{lcc} 
\toprule
  {\bf Pooling operator} & \macrof & \microf \\
\midrule
 Mean + Max Pooling & \textbf{78.3} & \textbf{70.6}  \\
 Mean Poolig & 76.9 & 68.1 \\
  Max Pooling  & \underline{77.6} &  \underline{69.5} \\ 
\texttt{[CLS]} Pooling & 77.1 & 69.5 \\
\bottomrule
\end{tabular}
\caption{Test results for various pooling operators with end-to-end tuning on SCOTUS for $\text{Longformer}_{\text{DA}}$.}
\label{tab:pool}
\end{table}

\section{Neural network Architecture}

Our model contains two linear layers with one activation layer and two batch normalization layers. 
We also compare the model without batch normalization layers. 
The comparison is made on the SCOTUS dataset using end-to-end fine-tuning.
One can see that removing batch normalization worsens performance.
\begin{table}[h]
\centering
\begin{tabular}{lcc}
\toprule
  {\bf Normalization }&  \macrof & \microf \\
    \midrule 
Batch Norm & {\bf 75.6} & {\bf 63.5}  \\
w/o Batch Norm & 72.5 & 63.1 \\
\bottomrule
\end{tabular}
\caption{F1 performance for ablation model without batch norm layers for end-to-end fine-tuning on SCOTUS.}
\label{tab:bn}
\end{table}

\end{document}